# Explaining Black-box Models for Biomedical Text Classification

Milad Moradi, Matthias Samwald

*Abstract*—In this paper, we propose a novel method named Biomedical Confident Itemsets Explanation (BioCIE), aiming at post-hoc explanation of black-box machine learning models for biomedical text classification. Using sources of domain knowledge and a confident itemset mining method, BioCIE discretizes the decision space of a black-box into smaller subspaces and extracts semantic relationships between the input text and class labels in different subspaces. Confident itemsets discover how biomedical concepts are related to class labels in the black-box's decision space. BioCIE uses the itemsets to approximate the black-box's behavior for individual predictions. Optimizing fidelity, interpretability, and coverage measures, BioCIE produces class-wise explanations that represent decision boundaries of the black-box. Results of evaluations on various biomedical text classification tasks and black-box models demonstrated that BioCIE can outperform perturbation-based and decision set methods in terms of producing concise, accurate, and interpretable explanations. BioCIE improved the fidelity of instance-wise and class-wise explanations by 11.6% and 7.5%, respectively. It also improved the interpretability of explanations by 8%. BioCIE can be effectively used to explain how a black-box biomedical text classification model semantically relates input texts to class labels. The source code and supplementary material are available at https://github.com/mmoradi-iut/BioCIE.

## I. INTRODUCTION

COMPLEX Artificial Intelligence (AI) models, such as deep neural networks, support vector machines, random forests, ensemble methods, etc. have been widely utilized in biomedical Natural Language Processing (NLP). Due to their immense capability of modeling lexical, syntactic, and semantic properties of natural language, these intricate methods have led NLP systems to achieve state-of-the-art results on a wide variety of biomedical and clinical text processing tasks [1-5]. However, these models inherently lack transparency and intelligibility, which restricts their application in real-world use cases. Users and developers often

Milad Moradi and Matthias Samwald are with Institute for Artificial Intelligence and Decision Support, Center for Medical Statistics, Informatics, and Intelligent Systems, Medical University of Vienna, 1090 Vienna, Austria (e-mails: {milad.moradivastegani, matthias.samwald}@meduniwien.ac.at).

need to know how a black-box AI system arrives at a particular decision, or how a set of inputs can lead to producing a specific output. Aiming at promoting transparency, fairness, and accountability of intelligent systems, eXplainable AI (XAI) methods try to disclose the inner working or decision logic of black-box AI models [6].

XAI models can be broadly divided into two categories of model-based and post-hoc methods [7]. The model-based explainability refers to designing and developing AI models whose inner working and decision making process is transparent to the user. Decision trees and linear regression models fall into this category. However, these simple models are not capable of learning nonlinear and complex data relationships; intricate black-box models, e.g. deep neural networks, are needed. Therefore, post-hoc explainability comes into play to show how a black-box AI model makes particular decisions or which relationships it has learned from the data. Post-hoc XAI methods can be of high importance in the biomedical NLP domain, since most of the recent successes have been achieved through utilizing complicated AI models [1, 2, 8]. Revealing how a NLP system relates a set of inputs to a particular output, a post-hoc XAI method can help the user discover the black-box's decision logic, biases in the model or data, and errors that the model is prone to.

In this paper, we propose a post-hoc explanation method named Biomedical Confident Itemsets Explanation (BioCIE) aiming at explaining black-box AI models used in biomedical text classification. Our goal is to tackle explainability issues imposed by perturbation-based and decision set explanators in the biomedical NLP domain. Perturbation-based methods approximate the behavior of a black-box by randomly changing the input and measuring the change in the output [9]. The problem is that the explanation is built based on random feature values that may never appear in the real inputs. Furthermore, random perturbation of textual inputs leads to meaningless data instances that may result in misleading explanations. Decision set explanation methods use frequent itemsets, i.e. feature values that frequently appear in the inputs, to create if-then rules that show how specific inputs result in producing an outcome [10]. The problem is that frequent words cannot properly define subspaces and decision logics within a classification model. An effective approach is needed to discover strong relationships between the inputs and black-box's predictions, no matter how frequent the input words are.

Addressing the above problems, BioCIE applies a confident itemset mining algorithm on real inputs to discretize the black-box's whole decision space into smaller subspaces. The explanator maps textual samples into biomedical concepts in order to capture semantic relationships between the inputs and the black-box's predictions. It extracts confident itemsets that discover strong relationships between concepts and class labels within each subspace. BioCIE uses confident itemsets to produce high-quality instance-wise explanations that accurately approximate the black-box's behavior for individual predictions. Utilizing an optimization procedure, BioCIE optimizes descriptive accuracy and interpretability measures on confident itemsets; it constructs class-wise explanations that accurately approximate decision boundaries of the black-box classifier.

We investigated the ability of BioCIE in producing accurate and interpretable explanations for predictions of three black-box models on various biomedical text classification tasks. The results showed that our BioCIE method outperforms perturbation-based and decision set explanators in terms of descriptive accuracy and interpretability of the explanations. BioCIE can be effectively used to explain predictions and approximate decision boundaries of black-box models in the biomedical text classification domain.

## II. RELATED WORK

With the widespread adoption of deep neural networks and other intricate computational models in the biomedical NLP and text processing fields [1, 8, 11], there has been a growing demand for XAI systems that help users understand how an opaque NLP model makes decisions or which relationships it has learned. So far, a wide variety of XAI techniques have been developed, each one aims at addressing challenges associated with a particular set of black-box models [6, 7]. Different classifications of XAI methods are proposed based on the explainability problem at hand, the type of explanator, the type of black-box model that is explained, and the modality of input data [12].

Based on the modelling stage in which explainability considerations come into play, XAI methods can be either model-based or post-hoc, which were already described in Section I. Post-hoc explainability can be further divided into local and global approaches [7]. A local XAI model provides an explanation for a single data record, while a global model tries to explain a whole predictive model learned by a black-box. The BioCIE method proposed in this paper is a post-hoc one, aiming at producing instance-wise and class-wise explanations. The former is used to approximate the local behavior of a black-box; the latter explains decision boundaries of a black-box within different decision subspaces characterized by class labels.

Explanations can be represented in the form of decision trees [13], decision rules [10], feature importance scores [9], partial dependence plots [14], prototypes [15], or other interpretable representations. Our BioCIE method provides explanations in the form of confident itemsets, i.e. a set of one or more biomedical concepts that are highly related to a class label, along with a confidence score that refers to the strength of the relationship between the concepts and the class label. In this way, interrelations between concepts in different parts of the decision space can be discovered. Moreover, the user is provided with a quantification of the strength of association between concepts and class labels. This helps to examine whether the underlying classifier is biased towards specific concepts when it produces an outcome.

An explanator may be designed to open only a particular type of black-box, e.g. tree ensembles [16], deep neural networks [17, 18], support vector machines [19], etc. On the other hand, a model-agnostic explanator is not tied to a specific class of black-box [9, 13, 20]. Our BioCIE method is a model-agnostic explanator; it receives a set of samples and respective predictions made by a black-box, and discovers how the classifier relates inputs to outputs in terms of semantics behind the text. This can help to investigate the ability of biomedical text classification systems in capturing semantic relationships between inputs and class labels, regardless of the underlying black-box.

Explanation methods may differ with regard to the data modalities they can handle. Some methods can be only applied to tabular [16], image [21], or text data [22], while some explanators were designed to work with various data modalities [9, 13]. Since the BioCIE explanation method utilizes biomedical domain knowledge to discover semantic relationships between input texts and predictions, it is considered a domain-specific explanator aiming at explaining black-box models on biomedical text classification tasks.

So far, few XAI studies in the biomedical domain have been devoted to textual data. Gao et al. [22] focused on explaining forest-based models for sentence classification in online health forums. Predicted samples were projected to an interpretable feature space in which sentences are represented using labeled sequential patterns, ontology-based, heuristic and sentence-based features. Decision rules were extracted from the new feature space to explain black-box predictions. Gehrmann et al. [23] defined a phrase-saliency metric to measure how much a phrase contributes to a prediction of a convolutional neural network for patient phenotyping from clinical narratives. In contrast to these works, our explanation method is designed to be model-agnostic. We also evaluate the efficacy of BioCIE for explaining various biomedical text classification tasks, instead of focusing on only one task.

## III. BioCIE EXPLANATION METHOD

Fig. 1 illustrates the overall architecture of our BioCIE explanation method. In this section, we first give a formulation of the problem, then every step of BioCIE is described in detail.

### A. Problem formulation

Let $f : X \rightarrow C$ be a black-box classifier, $X=\{X_1, \ldots, X_M\}$ be the set of input samples, and $C=\{C_1, \ldots, C_Q\}$ be the set of class labels in the classification problem. Given a set of class labels $Y=\{Y_1, \ldots, Y_M\}$ predicted by $f$, such that $Y_m$ is the class label assigned to instance $X_m$, the goal is to construct an instance-wise explanation $E_m$ for every individual prediction made by the classifier and a class-wise explanation $CE_q$ for every class $C_q$.

Confident itemset generation is the main building block of

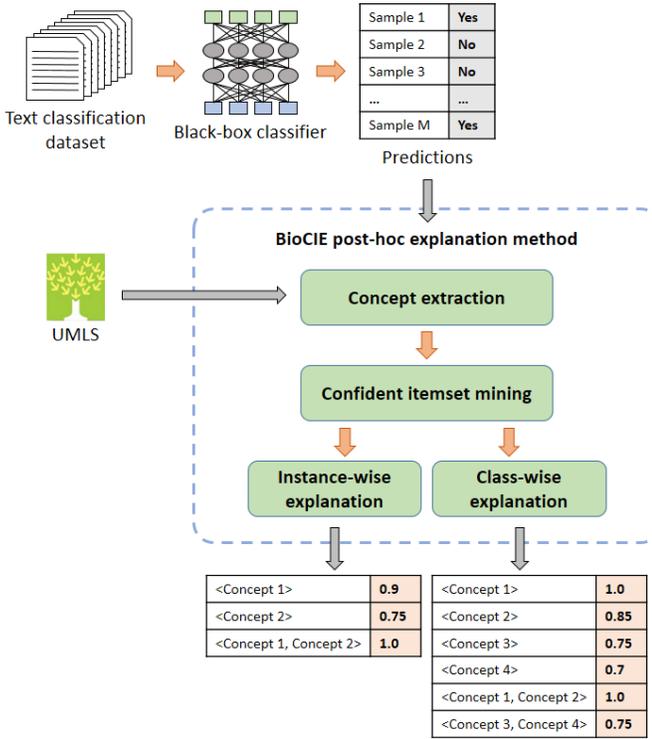

Fig. 1. The overall architecture of our BioCIE explanation method.

Fig. 2. A text sample from the BioText dataset and the extracted biomedical concepts in the first step of the BioCIE explanation method.

**Class: TREATMENT_FOR_DISEASE**

{Aspirin therapy} Confidence: 1.0
{Excision} Confidence: 1.0
{Induction Chemotherapy} Confidence: 1.0
{Combination Drug Therapy} Confidence: 1.0
{Pharmaceutical Solutions} Confidence: 0.91
{Surgical approach} Confidence: 0.85
{Persistent Disease} Confidence: 0.85
{Curative Surgery, Immunotherapy} Confidence: 1.0
{removal technique, Excision} Confidence: 1.0
{Oral Medicine, Antidiabetics} Confidence: 0.89
{Steroid therapy, Persistent Disease} Confidence: 0.81
{Antidiabetics, Classic Histology, Biguanides} Confidence: 1.0

Fig. 3. Some confident itemsets extracted for the class *TREATMENT_FOR_DISEASE* in the BioText dataset. In this example, the value of *min_conf* was set to 0.8.

BioCIE. An *Itemset* contains one or more biomedical concepts that appear in the samples belonging to a particular class $C_q$. A confidence property is computed for *Itemset* in a subspace characterized by $C_q$, as follows:

$$Confidence(Itemset, C_q) = P(Itemset | C_q) / P(Itemset) \quad (1)$$

where $P(Itemset)$ is the probability of observing *Itemset* in the set of all input samples, and $P(Itemset | C_q)$ is the probability of observing *Itemset* in samples belonging to class $C_q$. A confident $K$-itemset is an itemset that contains $K$ unique concepts and satisfies two criteria: 1) the confidence property of the itemset is equal to or greater than a threshold *min_conf*, and 2) every subset of the itemset is a confident itemset on its own.

An instance-wise explanation is represented as $E_m = \{<Itemset_1>, \ldots, <Itemset_P> \mid Y'_m\}$ such that $Y'_m$ is the class label assigned by the BioCIE explanation method to sample $X_m$, and $<Itemset_P>$ is a confident itemset containing one or more biomedical concepts that appear in $X_m$ and are highly correlated with class label $Y'_m$. A class-wise explanation is represented as $CE_q = \{<Itemset_1>, \ldots, <Itemset_R>\}$ such that $<Itemset_r>$ is a confident itemset containing one or more biomedical concepts that are highly discriminative in the decision subspace characterized by the class $C_q$.

### B. Concept extraction

Given a set of samples $X = \{X_1, \ldots, X_M\}$ such that $X_m$ contains a number of words, the explanation method begins by mapping the samples to the biomedical concepts contained in the Unified Medical Language System (UMLS) [24]. The UMLS is a large biomedical knowledge source that integrates more than 100 vocabularies, classification systems, and ontologies in the biomedical domain. It contains three main components; 1) the *Metathesaurus*, a lexicon of millions of concepts in life sciences and medicine, and their relationships; 2) the *Specialist Lexicon* that stores syntactic, morphological and orthographic information of the general English and biomedical vocabularies; 3) the *Semantic Network* that defines a set of semantic relations between the concepts and categorizes them into various semantic types.

We used the MetaMap tool [25] to extract biomedical concepts from the samples. MetaMap utilizes NLP techniques to match phrases in the input text with the UMLS concepts. In this study, we used the 2016 version of MetaMap, along with the '2018AA' release of UMLS as the knowledge base. Fig. 2 shows a sample from the BioText dataset and the extracted concepts in the first step of the BioCIE explanation method.

### C. Confident itemset mining

In this step, the decision space is divided into subspaces such that every subspace is characterized by a class $C_q$; biomedical concepts that are highly related to $C_q$ and correlations between concepts in the respective subspace are extracted. To this end, we utilize the confident itemset mining algorithm proposed in [26].

The confident itemset mining procedure is carried out in an iterative manner, where a set of confident $K$-itemsets are extracted for every class $C_q$ in the $K_{th}$ iteration. The confident itemset mining process continues until $K$ reaches a predefined value or no confident itemsets are extracted in the latest iteration. At the end of this step, $CI = \{CI_1, \ldots, CI_Q\}$ represents the set of all confident itemsets extracted for the whole

**Instance:** <u>combination chemotherapy</u> is the cornerstone of treatment that confers a meaningful survival benefit for patients with <u>small-cell lung cancer</u>.

**Real calls:** TREAT_FOR_DIS      **Prediction:** TREAT_FOR_DIS

*Explanation by BioCIE*

| *Itemset* | *Confidence* |
|---|---|
| {Combination Chemotherapy} | 1.0 |
| {Small cell carcinoma of lung} | 0.96 |
| {Small cell carcinoma of lung, Combination Chemotherapy} | 1.0 |

**Instance:** Patients were randomly assigned either <u>roxithromycin 150 mg</u> orally twice a day ( n = 102 ) or <u>placebo</u> orally twice a day ( n = 100 ).

**Real calls:** TREAT_ONLY      **Prediction:** TREAT_ONLY

*Explanation by BioCIE*

| *Itemset* | *Confidence* |
|---|---|
| {Roxithromycin 150 MG} | 1.0 |
| {Placebos} | 0.81 |
| {Roxithromycin 150 MG, Placebos} | 1.0 |

Fig. 4. Two instance-wise explanations produced by BioCIE for predictions of a black-box classifier on two samples from the BioText dataset.

decision space; every $CI_q$ represents confident itemsets extracted for class $C_q$. Fig. 3 shows some confident itemsets extracted for the class *TREATMENT_FOR_DISEASE* in the BioText dataset. In this example, seven 1-itemsets, four 2-itemsets, and one 3-itemsets are represented.

### D. Instance-wise explanation

The BioCIE explanation method uses the confident itemsets extracted in the previous step to approximate the local behavior of the black-box $f$. Given an instance $X_m$, a set of confident itemsets $CI=\{CI_1, …, CI_Q\}$ such that $CI_q=\{Itemset_1, …, Itemset_J\}$ represents the set of confident itemsets extracted for class $C_q$, an instance-wise explanation $E_m=\{<Itemset_1>, …, <Itemset_P> \mid Y'_m\}$ is produced that approximates the local behavior of $f$ for instance $X_m$. The set $<Itemset_1>, …, <Itemset_P>$ is constructed by searching over every $CI_q \in CI$ and extracting those confident itemsets that appear in $X_m$. A Confidence Score (CS) is computed for $E_m$ within every class $C_q$ such that there is at least one $Itemset_p \in CI_q$ and $Itemset_p$ appears in $X_m$, as follows:

$$CS(E_m, C_q) = \sum_{p=1}^{P} Confidence(Itemset_p, C_q) \qquad (2)$$

where $CS(E_m, C_q)$ is the confidence score of explanation $E_m$ within class $C_q$, and $Confidence(Itemset_p, C_q)$ is the confidence value of $Itemset_p$ in class $C_q$.

The explanation method selects the class label that obtained the highest confidence score and assigns it to $Y'_m$ as the local approximation of the black-box classifier $f$. Fig. 4 shows two instance-wise explanations produced by BioCIE for predictions of a black-box classifier on two samples from the BioText dataset. In these examples, the itemsets represent those biomedical concepts that appear in the input and are highly related to the predicted class label in the decision space learned by the target black-box classifier. It can be interpreted that the black-box in Fig. 4 has learned to classify a sample as

'*TREATMENT_FOR_DISEASE*' if the input contains disease and treatment entities such as '*small-cell lung cancer*' and '*combination chemotherapy*'. However, another black-box may learn to predict the same class label if the word '*treatment*' appears in the input, or it may even learn wrong relations between the input text and class labels. The BioCIE explanation method aims to reflect these relations learned by the target black-box, not to be used for identifying named entities or as a relation classifier on its own.

### E. Class-wise explanation

A class-wise explanation is an approximation of the behavior of the target black-box that shows what concepts and relationships between concepts lead the black box to produce a specific outcome. The set of confident itemsets $CI_q$ contains those concepts that are highly associated with class $C_q$. However, it may not be efficient to represent all itemsets in $CI_q$ as the class-wise explanation of $C_q$ since the large number of itemsets reduces the interpretability and understandability of the explanation. In order to deal with this problem, we define six properties that quantifies the fidelity, interpretability, and coverage of a class-wise explanation, then select an optimal subset of itemsets in $CI_q$ that optimizes the six properties.

Given a set of confident itemsets $CI_q=\{Itemset_1, …, Itemset_J\}$ extracted for class $C_q$, and a class-wise explanation $CE_q=\{<Itemset_1>, …, <Itemset_R>\}$, a *Fidelity* property is defined for $CE_q$ to quantify how accurately the explanation can mimic the behavior of the black-box in the respective subspace, as follows:

$$Fidelity(CE_q)=$$
$$[\sum_{m=1}^{M} X_m \epsilon X \mid f(X_m)=C_q \text{ and } f(X_m)=BioCIE(X_m)]/M_q \qquad (3)$$

where $BioCIE(X_m)$ is the class label assigned to instance $X_m$ by the BioCIE explanator, and $M_q$ is the total number of instances in $X$ classified into class $C_q$ by black-box $f$.

*Size*, *NumConcepts*, *MaxLength*, and *ItemsetOverlap* are four properties that quantify the interpretability. The size of explanation $CE_q$ is defined as the total number of itemsets in $CE_q$:

$$Size(CE_q) = \sum_{r=1}^{R} Itemset_r \qquad (4)$$

*NumConcepts* is computed by summing up the size of confident itemsets in $CE_q$:

$$NumConcepts(CE_q)=\sum_{r=1}^{R} Size(Itemset_r) \qquad (5)$$

where $Size(Itemset_r)$ is the total number of concepts that appear in $Itemset_r$.

*MaxLength* is the maximum size of a confident itemset in $CE_q$:

$$MaxLength(CE_q) = max\ Size(Itemset_r) \qquad (6)$$

*ItemsetOverlap (IO)* is defined as the total number of confident itemset pairs in $CE_q$ that have at least one concept in

common:

$$IO(CE_q)=\sum_{r=1}^{R}\ \sum_{h=1}^{H}\ IS_r,\ IS_h\ \epsilon\ CE_q\ |$$

$$Overlap(IS_r,\ IS_h)=True \qquad (7)$$

where $Overlap(IS_r,\ IS_h)$ is True if itemsets $IS_r$ and $IS_h$ have at least one concept in common.

A *Coverage* property is defined for class-wise explanation $CE_q$ as the total number of instances that were classified into class $C_q$ and are covered by a confident itemset in $CE_q$, as follows:

$$Coverage(CE_q)=$$

$$\sum_{m=1}^{M}\ X_m\epsilon X\ |\ f(X_m)=C_q\ \textbf{and}\ Cover(CE_q,\ X_m)=True \qquad (8)$$

where $Cover(CE_q,\ X_m)$ is *True* if at least one *Itemset$_r$* $\epsilon$ $CE_q$ appears in instance $X_m$.

Non-negative reward functions are defined for the six properties. When lower values are preferred for a property, the computed value is subtracted from the upper bound value. The reward functions are presented in TABLE I. The fidelity, interpretability, and coverage properties are jointly optimized using the six reward functions and the following objective:

$$max\ _{CE_q\subseteq CIq}\ \sum_{i=1}^{6}\ w_i f_i(CE_q) \qquad (9)$$

where $w_1,\ ...,\ w_6$ are positive weights that control the relative importance of the reward functions and are selected through cross-validation. The following constraints apply to the optimization objective:

$$Size(CE_q)\le\theta_1$$

$$NumItems(CE_q)\le\theta_2 \qquad (10)$$

$$MaxLength(CE_q)\le\theta_3$$

where $\theta_1$, $\theta_2$, and $\theta_3$ are specified by the user.

As already proven by Lee et al. [27] and Lakkaraju et al. [10], the objective given by (9) is submodular, non-monotone, non-negative, and non-normal. We use the optimization method proposed by Lee et al. [27] since it guarantees an optimal solution, relying on approximate local search. TABLE II gives a pseudo-code of the optimization algorithm utilized for selecting an optimal subset of confident itemsets to produce a class-wise explanation. Finally, those biomedical concepts that are highly associated with a class label and strong correlations between concepts in the respective subspace are represented in the form of a class-wise explanation.

Please note that our post-hoc explanation method is not intended to be used as a classification model; the goal is to produce explanations that reflect both correct and wrong decisions made by the underlying black-box. That is the reason why we do not discriminate between correct and wrong classifications produced by the black-box in equations (3) and (8).

## IV. EVALUATION METHOD

In order to evaluate the ability of our BioCIE explanation method in producing accurate and interpretable explanations for black-box classifiers, we implemented classification systems based on three AI models, i.e. BioBERT, LSTM, and SVM.

**BioBERT** [3] is a version of the well-known Bidirectional Encoder Representations from Transformers (BERT) [28] that was pretrained on massive corpora of biomedical text. We used the BioBERT-base version 1.0 pretrained on the PubMed and PMC corpora along with the default hyperparametr settings. **Long Short-Term Memory (LSTM)** [29] is a variation of recurrent neural networks that can effectively handle long-term dependencies in sequence data such as text. We used the *keras* library in *python* to implement the LSTM text classifier. We also used 100-dimensional embeddings as the input, the *softmax* activation in the output layer, *binary cross entropy* as the loss function, and the *Adam* optimizer with the default settings. **Support Vector Machines (SVM)** [30] separate samples of two classes in a high-dimensional vector space by maximizing the distance between the classes' margins and the separator hyperplane. Using the *Scikit-learn* library [31], we implemented a SVM classifier with *RBF* kernel, and *tf-idf* weights as the input features.

We conducted experiments on three biomedical text classification tasks. **BioText** is a dataset for disease-treatment information extraction [32], which can be also used as a text classification dataset. It contains more than 3,500 text samples classified into one of the eight classes *DISONLY*, *TREATONLY*, *TREAT_FOR_DIS*, *PREVENT*, *SIDE_EFF*, *TREAT_NO_FOR_DIS*, *TO_SEE*, *NONE*, and *VAGUE*, which specify the type of semantic relationship between disease and treatment entities appearing in the text. The class label *TO_SEE* refers to those samples that contain more than one type of relationship between diseases and treatments, in order to be further examined by an expert. The black-boxes were not trained to identify and extract disease and treatment named entities; they were trained to classify each sample into one of the eight classes. The **AIMed** dataset [33] includes more than 2,000 text segments from PubMed abstracts in which protein-protein interactions are annotated. We used this dataset for text classification such that the annotations were removed and a class label added to each sample. The class label specified whether the sample conveys a protein-protein interaction or not. Again, the black-boxes were trained for the text classification task, not for the named entity or relation extraction. The **Hereditary Diseases (HD)** dataset [34] contains more than 550 biomedical articles classified into one of the 26 hereditary diseases. We split each dataset into separate training and test sets with a ratio of 90:10. The black-box classifiers were trained on the training sets, and the explanations were produced for the outcomes of the black-boxes on the test sets. TABLE III presents the predictive accuracy scores obtained by the black-box classifiers on the three biomedical text classification datasets.

We compared our BioCIE explanation method against four baselines, i.e. LIME, MUSE, Greedy, and Random. **LIME** [9] is a publicly-available explanator that relies on random perturbations and local linear model approximation. **MUSE** [10] utilizes frequent itemsets to generate predicates that define subspace descriptors and decision logics of a model in the form of if-then rules. We implemented MUSE according to the method described in the respective paper. Words were used as itemsets; presence of high-frequent words (frequent itemsets) were checked by if-then rules. The **Greedy** baseline was implemented based on the heuristic proposed by Martens et al. [35]. It greedily selects $N$ important words that have the highest impact on choosing a class label. The **Random** baseline randomly selects $N$ words form the samples assigned to a class label as the explanation for the respective class.

## V. RESULTS AND DISCUSSION

In this section, we present and discuss the experimental results that evaluate our BioCIE method and the other explanators in terms of fidelity and interpretability of the explanations produced for outcomes of the black-box classifiers.

### A. Fidelity

Descriptive accuracy (or fidelity) refers to the ability of an explanator in imitating the behavior of a black-box in terms of assigning class labels to data instances [7]. It has been widely used in evaluating the efficacy of XAI systems [9, 36, 37]. The higher the fidelity of the explanations, the more accurate the approximation of the black-box's behavior.

#### 1) Instance-wise explanation

For instance-wise explanations produced by LIME, we set the number of feature importance scores to 10. MUSE produced instance-wise explanations by extracting the decision rules that apply to individual text samples. TABLE IV presents the fidelity scores obtained by BioCIE and the other methods for instance-wise explanation. The explanation methods were experimented on every black-box classifier and every classification task.

TABLE III
THE ACCURACY SCORES OBTAINED BY THE BLACK-BOX CLASSIFIERS ON THE THREE BIOMEDICAL TEXT CLASSIFICATION DATASETS

| Black-box classifier | Dataset | | |
|---|---|---|---|
| | BioText | AIMed | HD |
| BioBERT | 0.917 | 0.931 | 0.908 |
| LSTM | 0.892 | 0.904 | 0.875 |
| SVM | 0.835 | 0.810 | 0.806 |

TABLE IV
THE FIDELITY SCORES OBTAINED BY THE EXPLANATION METHODS FOR INSTANCE-WISE EXPLANATION

| Explanator | Black-box | Dataset | | |
|---|---|---|---|---|
| | | BioText | AIMed | HD |
| LIME | BioBERT | 0.849 | 0.852 | 0.830 |
| | LSTM | 0.836 | 0.855 | 0.828 |
| | SVM | 0.807 | 0.834 | 0.793 |
| MUSE | BioBERT | 0.775 | 0.788 | 0.755 |
| | LSTM | 0.771 | 0.783 | 0.746 |
| | SVM | 0.758 | 0.770 | 0.761 |
| Greedy | BioBERT | 0.670 | 0.653 | 0.659 |
| | LSTM | 0.683 | 0.664 | 0.641 |
| | SVM | 0.691 | 0.677 | 0.665 |
| Random | BioBERT | 0.482 | 0.495 | 0.441 |
| | LSTM | 0.495 | 0.479 | 0.452 |
| | SVM | 0.456 | 0.494 | 0.476 |
| BioCIE | BioBERT | 0.917 | 0.925 | 0.889 |
| | LSTM | 0.908 | <u>0.931</u> | 0.915 |
| | SVM | <u>0.922</u> | 0.910 | <u>0.921</u> |

TABLE V
THE FIDELITY SCORES OBTAINED BY THE EXPLANATION METHODS FOR CLASS-WISE EXPLANATION

| Explanator | Dataset | | |
|---|---|---|---|
| | BioText | AIMed | HD |
| LIME-10 | 0.592 | 0.604 | 0.588 |
| LIME-20 | 0.696 | 0.700 | 0.684 |
| LIME-30 | 0.783 | 0.788 | 0.775 |
| LIME-40 | 0.810 | 0.819 | 0.807 |
| LIME-50 | 0.831 | 0.843 | 0.829 |
| MUSE | 0.752 | 0.771 | 0.732 |
| Greedy | 0.758 | 0.749 | 0.718 |
| Random | 0.493 | 0.514 | 0.517 |
| BioCIE | <u>0.885</u> | <u>0.907</u> | <u>0.881</u> |

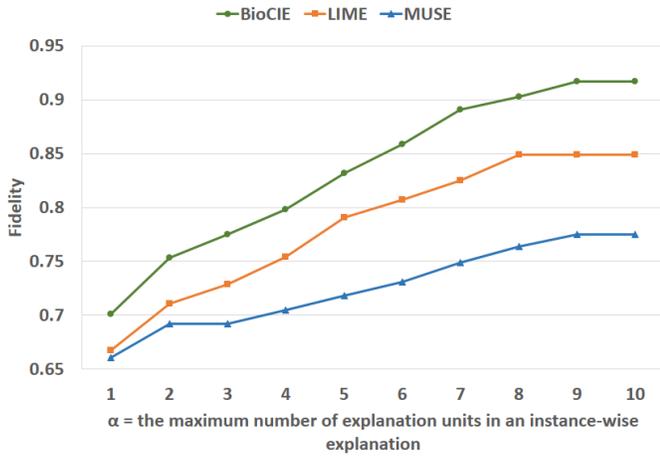

Fig. 5. The maximum number of explanation units in an instance-wise explanation against the descriptive accuracy results. (Experiment 1)

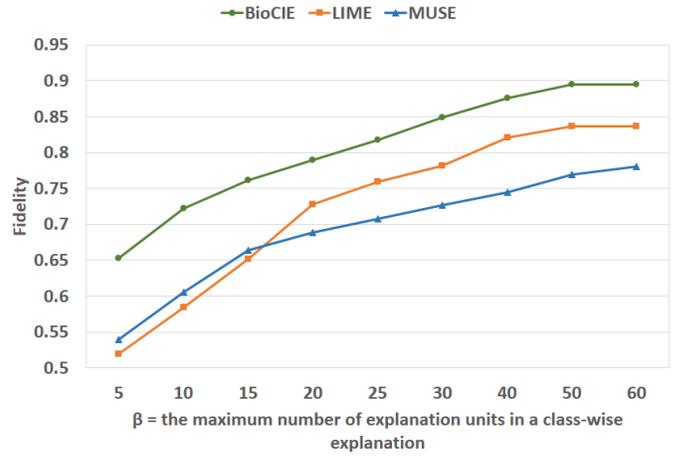

Fig. 6. The maximum number of explanation units in a class-wise explanation against the descriptive accuracy results. (Experiment 2)

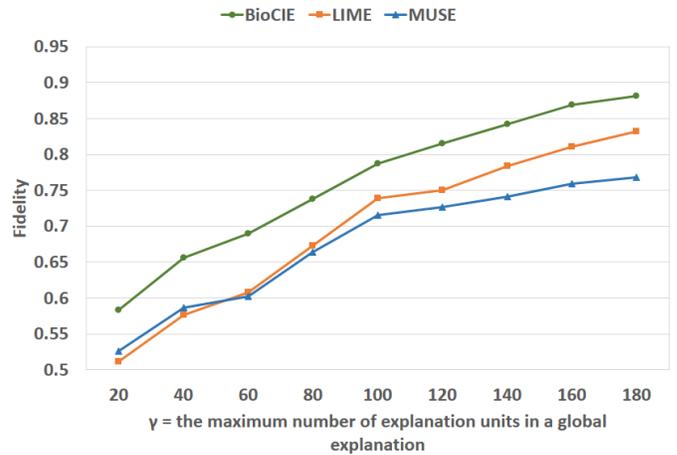

Fig. 7. The maximum number of explanation units in a global explanation against the descriptive accuracy results. (Experiment 3)

As the results show, BioCIE outperforms the other explanators on all the three datasets and three black-box models. This demonstrates that the confident itemsets can capture relationships between the inputs and the black-boxes' predictions more accurately than the perturbation and decision set methods. Mapping the input text to biomedical concepts, BioCIE discovers how the black-box semantically associates an input with a class label. In this way, the explanator does not need to see a set of particular words in the input in order to approximate a specific class label. It can mimic the black-box's behavior with respect to the biomedical semantics conveyed by the input.

Examining the explanations produced by MUSE, we observed that those words that frequently appear in the samples belonging to a class have the most impact on assigning the respective class label by the explanator. In fact, high-frequent words are used to discriminate between the classes. We also observed that LIME sometimes uses unrelated words when it approximates a class label, or assigns higher importance scores to unimportant words. This may be caused by the perturbation method that generates random instances that do not reflect the real distribution of inputs.

### 2) Class-wise explanation

We evaluated class-wise explanations to assess the ability of the methods in approximating decision boundaries of the black-boxes. Given a class label, a class-wise explanation was produced for LIME by collecting $N$ words that were assigned the highest feature scores in the instance-wise explanations belonging to the class. We tested LIME's class-wise explanations for $N$=10, 20, 30, 40, and 50; each setting is referred to as LIME-$N$. Given a class label, a class-wise explanation was produced for MUSE by collecting those decision rules that lead to the same class. Each explanator approximated the decision boundaries of the black-box classifiers by assigning class labels to the samples according to the class-wise explanations. TABLE V presents the fidelity scores obtained by the explanators for class-wise explanation. For brevity reasons, the average of fidelity scores of the explanations generated for the outcome of the three black-box classifiers is reported for each explanator.

As the results show, BioCIE can produce more accurate class-wise explanations than the other methods. This superiority in accurate approximation of the black-box's decision boundaries is attributed to two elements: 1) the confident itemsets that accurately capture semantic relationships between inputs and predictions in a given subspace, and 2) the optimization process that helps to extract an optimal subset of itemsets that best approximate the black-box's behavior in the given subspace.

Although visualizing attention weights has been helpful to understand how important a word is when computing the next representation for other words of an input sequence in a transformer model [38], some studies have shown there is no correlation between attention weights learned by BERT (and its variants such as BioBERT) and feature importance scores produced by gradient-based and leave-one-out methods [39]. Therefore, it has been suggested not to treat attention as justification for a transformer model's decisions [40]. However, the results show that BioCIE can effectively address the explainability issue of transformer models; it produced accurate approximations of the behavior of BioBERT.

### B. Interpretability

Interpretability refers to how easily an explanation can be understood or interpreted. It can be evaluated by different criteria and measures, depending on the task at hand and the representation used to show explanations. Regarding the assumption that smaller explanations are more interpretable [37], we conducted a set of experiments to investigate the interpretability of the explanators in terms of the trade-off between descriptive accuracy and interpretability measures. For brevity reasons, we report the average of fidelity scores obtained for the explanations produced on outcomes of the **BioBERT** classifier on the three text classification datasets. We observed similar results when the other black-box classifiers were used.

*Experiment 1*: Given a parameter $\alpha$ that specifies the maximum number of explanation units in an instance-wise explanation, we assessed the ability of the explanators in producing as small and accurate explanations as possible. An explanation unit refers to an itemset in BioCIE, a feature and the respective importance score in LIME, and an if-then rule in MUSE. We varied the value of $\alpha$ and measured the descriptive accuracy of the instance-wise explanations. Fig. 5 shows the results.

*Experiment 2*: Given a parameter $\beta$ that specifies the maximum number of explanation units in a class-wise explanation, we assessed the ability of the explanators in producing small and accurate explanations. We varied the value of $\beta$ and measured the descriptive accuracy of the class-wise explanations. Fig. 6 shows the results.

*Experiment 3*: Given a parameter $\gamma$ that specifies the maximum number of explanation units in a global explanation, we assessed the ability of the explanators in producing small and accurate global explanations. A global explanation was generated by selecting an optimal subset of $\gamma$ explanation units that optimize two criteria: 1) as many text samples as possible should be covered by the global explanation, and 2) as few text samples as possible should be covered by more than one explanation units that do not lead to the same class label. We varied the value of $\gamma$ and measured the descriptive accuracy of the global explanations. Fig. 7 shows the results.

As the results of all the three experiments show, our BioCIE explanation method achieved higher levels of descriptive accuracy (or fidelity) when it used smaller sets of explanation units compared with the other explanators, which demonstrates that BioCIE produces more interpretable explanations. For example, BioCIE used an average of 4, 22, and 110 explanation units to achieve a descriptive accuracy of 80% on instance-wise, class-wise, and global explanations, respectively. LIME needed an average of 5.5, 35, and 150 explanation units to reach a descriptive accuracy of 80% on instance-wise, class-wise, and global explanations, respectively. MUSE did not reach a descriptive accuracy of 80% in our experiments; it generally used more explanation units than the other explanators to hit the same descriptive accuracy score.

As can be observed from Fig. 5-7, increasing the size of explanations led to an increase in the fidelity of the explanations. However, this linear relationship between the fidelity and the size of explanations did not always exist, and there was no improvement in the fidelity (or there was a slight improvement) after the size of explanations reached a certain threshold. This suggests that adding more explanation units can improve the fidelity when the explanation is not large enough to approximate the black-box's behavior in different parts of the decision space. However, reaching a certain level of descriptive accuracy, larger explanations not only do not lead to higher fidelity, but also decrease the interpretability.

The confident itemset mining phase and the optimization procedure played key roles in generating concise, accurate, and interpretable explanations by BioCIE. The confident itemsets discovered semantic relationships between the input text and the class labels in different decision subspaces. The confidence property of itemsets enabled the explanator to assess the strength of semantic relationships in every subspace and produce a concise list of biomedical concepts that accurately approximate the target classifier's behavior when it predicted a specific class. This concept-based approach to relation extraction and black-box approximation efficiently reduced the size of instance-wise explanations and improved the descriptive accuracy. Optimizing fidelity, interpretability, and coverage properties effectively reduced the size of itemsets that defined the decision boundaries within a subspace, leading to smaller and more interpretable, yet accurate class-wise explanations. It is worth-mentioning that the confident itemsets explanation approach obtained higher scores than LIME and MUSE in subjective evaluation of usability and interpretability by users [26].

## VI. CONCLUSION

In this paper, we proposed the BioCIE method for post-hoc explanation of black-box machine learning models applied to biomedical text classification. The results of evaluations showed that BioCIE outperforms the perturbation-based and decision set explanators in terms of fidelity and interpretability of instance-wise and class-wise explanations. Summing up the results, we point out the following concluding remarks:

- Combining biomedical concepts and confident itemset mining is an effective approach to discovering semantic relations between inputs and outputs of a black-box biomedical text classification model.
- Concise, accurate, and interpretable explanations can be produced by optimizing fidelity and interpretability measures on subsets of confident itemsets.
- The BioCIE method can effectively reveal decision boundaries and approximate behavior of black-box text classification models in the biomedical domain.

So far, we have discussed how BioCIE can approximate and reveal decision boundaries of a target black-box. It can be also beneficial to discuss how BioCIE could help correct decisions of the black-box. If there are biases in the data, the black-box may learn wrong relations between biomedical concepts and class labels. In this case, confident itemsets can reflect these wrong relations and disclose the biases. Removing erroneous training samples and/or adding more training samples could be appropriate corrective strategies in this situation. The black-box model itself may also lead to wrong classifications. In this case, confident itemsets disclose limitations of the model by presenting erroneous relations between the input space and class-labels. This type of error could be caused by the model's inability to learn semantic, lexical, or even syntactical relations. Therefore, changing the model's architecture or hyperparameters, or even choosing other types of nonlinear predictive models could properly enhance the black-box's performance.

Future work includes applying BioCIE to other biomedical NLP tasks. This requires further customization of the itemset mining and optimization steps. One more idea for future work could be customizing BioCIE to produce task-specific explanations. For example, customizing BioCIE for the classification task of the BioText dataset, explanations could be represented as sentences such as '*[Procedure X] is a treatment for [Disease Y]*' or '*[Procedure X] is a preventive measure for [Disease Y]*'. In this way, the explanation shown in Fig. 4 can be represented as '*Combination Chemotherapy is a treatment for Small cell carcinoma of lung*'.